\documentclass[10pt,twocolumn,letterpaper]{article}

\usepackage[pagenumbers]{wacv} 

\usepackage{graphicx}
\usepackage{amsmath}
\usepackage{amssymb}
\usepackage{booktabs}
\usepackage{mathtools}
\usepackage{algpseudocode,algorithm}
\usepackage[dvipsnames]{xcolor}
\usepackage{adjustbox}
\usepackage{pifont}
\usepackage{amssymb}
\usepackage{multirow}
\usepackage{colortbl}
\usepackage{wrapfig}
\usepackage{subcaption}

\usepackage[pagebackref,breaklinks,colorlinks]{hyperref}

\usepackage[capitalize]{cleveref}
\crefname{section}{Sec.}{Secs.}
\Crefname{section}{Section}{Sections}
\Crefname{table}{Table}{Tables}
\crefname{table}{Tab.}{Tabs.}

\def\wacvPaperID{2039} 
\def\confName{WACV}
\def\confYear{2025}

\begin{document}

\title{Improving Accuracy and Generalization for Efficient Visual Tracking}

\author{Ram Zaveri \qquad Shivang Patel \qquad Yu Gu \qquad  Gianfranco Doretto\\ 
West Virginia University\\
Morgantown, WV 26506, USA\\
{\tt\small \{rz0012, sap00008, yugu, gidoretto\}@mix.wvu.edu}
}
\maketitle
\begin{abstract}
\vspace{-3mm}
Efficient visual trackers overfit to their training distributions and lack generalization abilities, resulting in them performing well on their respective in-distribution (ID) test sets and not as well on out-of-distribution (OOD) sequences, imposing limitations to their deployment in-the-wild under constrained resources. We introduce SiamABC, a highly efficient Siamese tracker that significantly improves tracking performance, even on OOD sequences. SiamABC takes advantage of new architectural designs in the way it bridges the dynamic variability of the target, and of new losses for training. Also, it directly addresses OOD tracking generalization by including a fast backward-free dynamic test-time adaptation method that continuously adapts the model according to the dynamic visual changes of the target. Our extensive experiments suggest that SiamABC shows remarkable performance gains in OOD sets while maintaining accurate performance on the ID benchmarks. SiamABC outperforms MixFormerV2-S by 7.6\% on the OOD AVisT benchmark while being 3x faster (100 FPS) on a CPU. Our code and models are available at \url{https://wvuvl.github.io/SiamABC/}.
\end{abstract}


\vspace{-6mm}
\section{Introduction}
\label{sec:intro}
\vspace{-1mm}

Tracking a single object, given the location at the first frame, has been an ongoing challenge in the vision community for decades. Most recent approaches provide reasonably good performance~\cite{gao2022aiatrack,he2023target,yan2021learning,wei2023autoregressive,chen2023seqtrack}, especially when benchmarked on \emph{in-distribution (ID)} datasets, \ie, on the testing portion of the same datasets used for training. However, they incur high computational costs and hardware constraints, making their deployment ``in-the-wild'' in mobile, autonomous, and IoT applications still challenging.

The best-performing Transformer-based trackers operate between 0.4 to 4 frames per second (FPS) on a CPU~\cite{chen2023seqtrack,wu2023dropmae}, which is considered ``slower than real-time'' in many applications. Siamese tracking approaches provide the highest speed. FEAR-XS~\cite{borsuk2022fear} can operate at 100 FPS on a CPU, whereas an efficient Transformer-based approach, MixFormerV2-S\cite{cui2024mixformerv2}, operates at 37 FPS on a CPU. Despite the significant progress on efficient trackers, they still need to catch up when tested on \emph{out-of-distribution (OOD)} datasets, \ie, those that were not used during training.

A recently proposed benchmark, AVisT\cite{noman2022avist}, involves tracking objects under extreme visibility conditions that are common in-the-wild but not in most current training sets. High-performing tracking approaches tend to struggle when tested on AVisT, showing very significant performance deterioration. For instance, MixFormerV2-S exhibits remarkable performance with AUC of 58.7\% on an in-distribution benchmark like GOT-10k\cite{Huang2021}; however, it struggles on the OOD benchmark AVisT, with an AUC of 39.6\%. Therefore, the trade-off between the need of computational resources and \emph{OOD generalization} abilities of visual trackers is still unsatisfactory for their deployment in-the-wild under resource constraints.  
\begin{figure}[t!]
    \centering
        \includegraphics[width=0.35\textwidth]{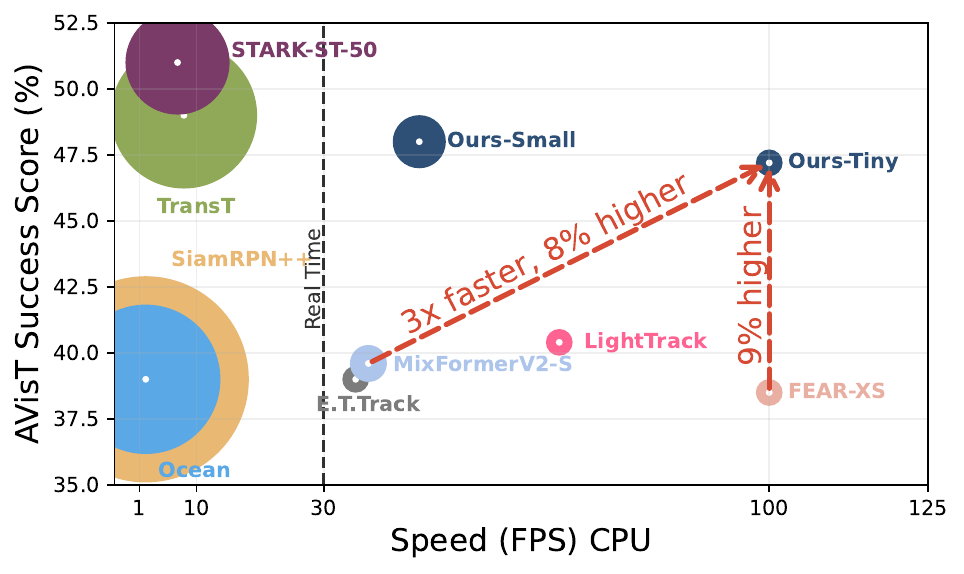}
    \vspace{-4mm}
    \caption{Comparison of our trackers with others on the AVisT~\cite{noman2022avist} dataset on a CPU. We show the success score (AUC) (vertical axis), speed (horizontal axis), and relative number of FLOPs (circles) of the trackers. Our trackers outperform other efficient trackers in terms of both speed and accuracy.}
    \label{fig:avist_comparision_fig}
    \vspace{-5mm}
\end{figure}

In this work we aim at significantly improving the trade-off mentioned above. We design a new Siamese tacker that preserves the high speed, reduced memory and computing requirements of the Siamese family while improving OOD generalization to near SOTA-level performance. From the architectural point of view, we make two key contributions. First, we better facilitate the visuo-temporal bridge between the static image template representing the target, and the search region image at current time. While~\cite{borsuk2022fear} popularized the use of a dual-template, which we also adopt, we introduce the use of a \emph{dual-search-region}. This will allow the tracker to stay anchored to the initial target representation while better latching onto its dynamic appearance variations. Second, we design a new learnable layer, the \emph{Fast Mixed Filtration}, that acts as an efficient filtration method for enhancing the relevant components of the combination of the representations forming the dual-template, as well as the dual-search-region. This is important because, given also the reduced representational capacity of smaller backbones used by efficient trackers, directly fusing the representations of the dual-template does not necessarily improve performance~\cite{borsuk2022fear}.

From the learning point of view we make two additional contributions. First, we introduce a new \emph{transitive relation loss} to help bridge the visuo-temporal similarities of the filtered representations of the dual-template and the dual-search-region, so that the relevant relational differences between them can be effectively leveraged for tracking purposes. Second, we more directly address the OOD generalization issue by tackling the dynamic distribution shifts while doing inference. As shown in many test-time adaptation (TTA) approaches for classification~\cite{wang2020tent,mirza2022norm, pan2018two, niu2022efficient, schneider2020improving, li2016revisiting}, shifts in Batch-Normalization (BN) statistics are majorly responsible for performance degradation under OOD testing. We introduce a \emph{dynamic TTA (DTTA)} approach specifically tailored to tracking. It is backward-free, thus lightweight computationally, and aims at dynamically updating the BN statistics while keeping them anchored to the source statistics. To the best of our knowledge this is the first work that uses TTA for single object visual tracking.

Combining the contributions above lead even our smallest and most efficient tracker, S-Tiny, to surpass relevant SOTA approaches on numerous benchmarks. Most notably, on the AVisT benchmark, S-Tiny achieves the AUC of 47.2\% while running at 100 FPS on a CPU, outperforming MixFormerV2-S by 7.6\% while being almost 3x faster. See \Cref{fig:avist_comparision_fig}. 
An extensive set of experiments with multiple datasets and other approaches shows additional compelling results in support of our method.


\vspace{-2mm}
\section{Related Works}
\vspace{-1mm}

\textbf{Efficient Tracking.}
 Practical applications require object trackers to be efficient as well as accurate. Siamese-based\cite{koch2015siamese,melekhov2016siamese}  trackers\cite{held2016learning, li2018high,zhu2018distractor, li2019siamrpn++, xu2020siamfc++, bertinetto2016fully, wang2019fast, chen2020siamese,yu2020deformable, yang2020siamatt} are efficient as they use a separate two stream feature-extraction framework. LightTrack\cite{yan2021lighttrack} and FEAR\cite{borsuk2022fear} introduced lightweight siamese-based trackers, however, lack accurate performance. Transformer-based approaches\cite{gao2022aiatrack, wei2023autoregressive,chen2023seqtrack, yan2021learning} show reasonable accuracy, however, they lack efficiency as they utilize computationally heavy attention layers. 
To alleviate that, E.T.Track~\cite{blatter2023efficient} incorporates an efficient Exemplar Transformer block on the prediction heads. While HCAT~\cite{chen2022efficient} uses multiple hierarchical cross-attention blocks with feature sparsification, HiT\cite{kang2023exploring}, instead, leverages a lightweight hierarchical transformer backbone to achieve improved accuracy and speed. MixformerV2~\cite{cui2024mixformerv2} uses distillation to reduce the number of FLOPs, whereas SMAT \cite{gopal2024separable} uses separable mixed attention to maintain the accuracy on their one-stream transformer networks. Since these are smaller networks, they have limited representational capacity and do not generalize well to OOD sets, making them less reliable for tracking ``in-the-wild", which we tackle in the proposed framework.

\textbf{Efficient Attention.}
\cite{mehta2022separable} proposed separable transformer blocks with convolutional layers to increase efficiency while maintaining accuracy. MobileViTv3~\cite{wadekar2022mobilevitv3} further replaced heavy transformer blocks with their CNN-ViT-based separable attention blocks, and showed considerable performance gain. Originally, CBAM~\cite{woo2018cbam}, DANet~\cite{fu2019dual}, and Polarized Self-Attention~\cite{liu2021polarized} explored attention in convolution by computing channel and spatial attentions separately. CBAM~\cite{woo2018cbam} and PSA~\cite{liu2021polarized} further incorporate a squeeze-and-excite framework~\cite{iandola2016squeezenet} to excite relevant features across the channels. The latter is a more powerful and efficient variant of CBAM.
This suggests that separability is inevitable for efficient attention blocks; therefore, in this work we propose a simplified fast convolution-based separable attention framework.

\textbf{Efficient Adaptation.}
To tackle the dynamic distribution shifts during inference, we focus on Test-Time Adaptation (TTA). A recent popular approach, CoTTA~\cite{wang2022continual} uses data augmentation at test time to generate pseudo-labels and performs distillation for the image classification task. TENT \cite{wang2020tent} and EATA \cite{niu2022efficient} use model's test-time entropy to update only the BN learnable parameters. DUA \cite{mirza2022norm}, Momentum \cite{schneider2020improving}, IN \cite{pan2018two}, and AdaBN \cite{li2016revisiting} use backward-free BN-statistics updates to perform adaptation with maximal efficiency while showing considerably good accuracy. Most TTA approaches experience performance deterioration when used under real-world online applications, except BN-adaptation approaches as they are efficient and reliable~\cite{alfarra2023revisiting}. Therefore, we propose an efficient instance-level BN update strategy that continuously adapts the model to follow the dynamic visual changes of the target.


\begin{figure*}[t]
    \hspace*{-10pt}
    \centering
    \includegraphics[width=0.8\textwidth]{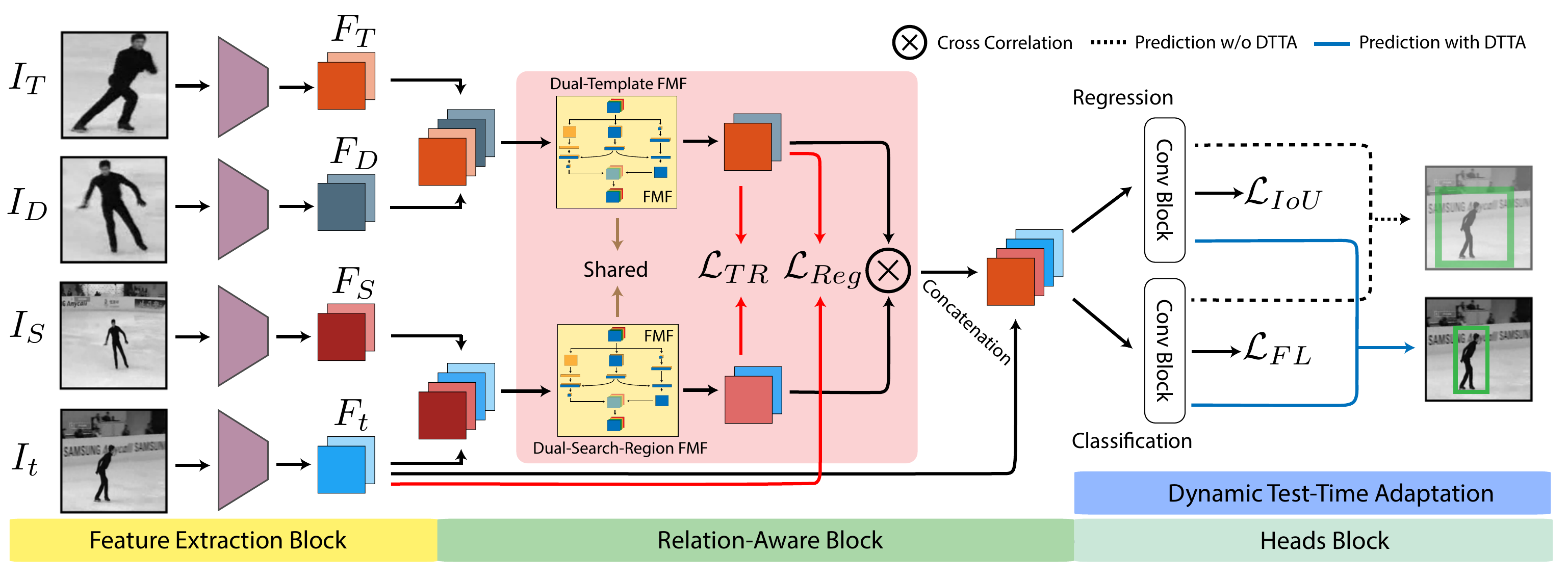}
    \vspace{-4mm}
     \caption{\textbf{Overall Architecture.} The Feature Extraction Block uses a readily available backbone to process the frames. The Relation-Aware Block exploits representational relations among the dual-template and dual-search-region through our losses, $\mathcal{L}_{TR}$ and $\mathcal{L}_{Reg}$, where dual-template and dual-search-region representations are obtained via our learnable FMF layer. The Heads Block learns lightweight convolution layers to infer the bounding box and the classification score through standard tracking losses, $\mathcal{L}_{IoU}$ and $\mathcal{L}_{FL}$ respectively. During inference, the tracker adapts to every instance through our Dynamic Test-Time Adaptation framework.}
     \label{fig:architecture}
     \vspace{-5mm}
  \end{figure*}
  
  \vspace{-2mm}
  \section{Methods}
\label{sec:methods}
  \vspace{-1mm}

\textbf{Overview.} We introduce a tracker that maintains four data sources. There is the \emph{static image template} $I_T$ that represents an object. The \emph{dynamic image template} $I_D$ instead, represents the object at a time $t - \Delta t$, where $t$ is the current time.
There is a \emph{search region image} $I_t$, where the object is presumed to be located at current time $t$. Unlike previous trackers, we also maintain a \emph{dynamic search region image} $I_{S}$, which is the image of the search region at time $t - \Delta t$ re-centered at the object position, i.e., it contains $I_D$ in the center. So, besides the \emph{dual-template}, $(I_T, I_D)$, our tracker incorporates temporal information also via the \emph{dual-search-region}, $(I_S, I_t)$. Specifically, the static template anchors the tracker at the object representation at time $t=0$ (A), the dynamic template and the dynamic search region represent time $t - \Delta t$ (B), and the search region represents time $t$ (C). The dual-template will lead to a boosted object representation that bridges the time gap $t - \Delta t$ (from A to B), while the dual-search-region will lead to a boosted search region representation that bridges the time gap $\Delta t$ (from B to C). Since we use a siamese architecture and leverage the relations between points in time A and B, and between points B and C, then blend them, we name our approach \emph{SiamABC}.

SiamABC utilizes a feature extraction backbone, a new \emph{Fast Mixed Filtration (FMT)} module, a \emph{Pixel-wise Cross-Correlation} module, and heads for classification scores and bounding box regressions. All the inputs, $I_T$, $I_D$, $I_{S}$, and $I_t$, go through the backbone $F(\cdot)$, giving us $F_T$, $F_D$, $F_S$, and $F_t$ respectively. Next, the pair $(F_T, F_D)$, and the pair $(F_S, F_t)$ go through the FMT module, producing $\Omega(F_T,F_D)$ and $\Omega(F_S,F_t)$, respectively.
Then, the Pixel-wise Cross-Correlation module, $CC(\cdot, \cdot)$, computes the correlation between $\Omega(F_T,F_D)$ and $\Omega(F_S,F_t)$, boosting their representational relations. The output of $CC(\cdot, \cdot)$ is further processed by the classification head, $CH(\cdot)$, and the bounding box regression head, $BH(\cdot)$, to produce the final tracking output. To learn representations that enable tracking by bridging from A to C, we introduce a new \emph{transitive relation} loss. Finally, to further adapt to dynamic shifts of the input distribution, which are typical when tracking is deployed ``in-the-wild'', on out-of-distribution data, we endow tracking, for the first time, with a \emph{dynamic backward-free test-time adaptation} approach. See \Cref{fig:architecture}.

\subsection{Architecture }\label{sec:architecture}

\textbf{Feature Extraction Block.}
For efficiency, we chose the first four layers of FBNetV2~\cite{wu2019fbnet} as our Tiny backbone, and the first three layers of ResNet-50~\cite{he2016deep} as our Small backbone, all pre-trained on ImageNet~\cite{deng2009imagenet}. Since the channel and spatial resolution of the backbones can differ, we use an additional convolutional filter (without activation) to match the channel resolution. The backbone takes in the input $x \in \mathbb{R}^{3 \times H \times W}$, where $H=W=128$ for $I_T$ and $I_D$, and $H=W=256$ for $I_t$ and $I_S$. The backbone processes the inputs in parallel, and the weights are shared. This functions in a siamese fashion as described in~\cite{bertinetto2016fully}.

\textbf{Relation-Aware Block.} The representations of the dual-template and the dual-search-region are first enhanced by the new Fast Mixed Filtration layer, and then correlated by the Pixel-wise Cross-Correlation module to support tracking.

\textsl{\textbf{Fast Mixed Filtration.}}
\begin{figure}[t]
    \centering
    \includegraphics[width=0.4\textwidth]{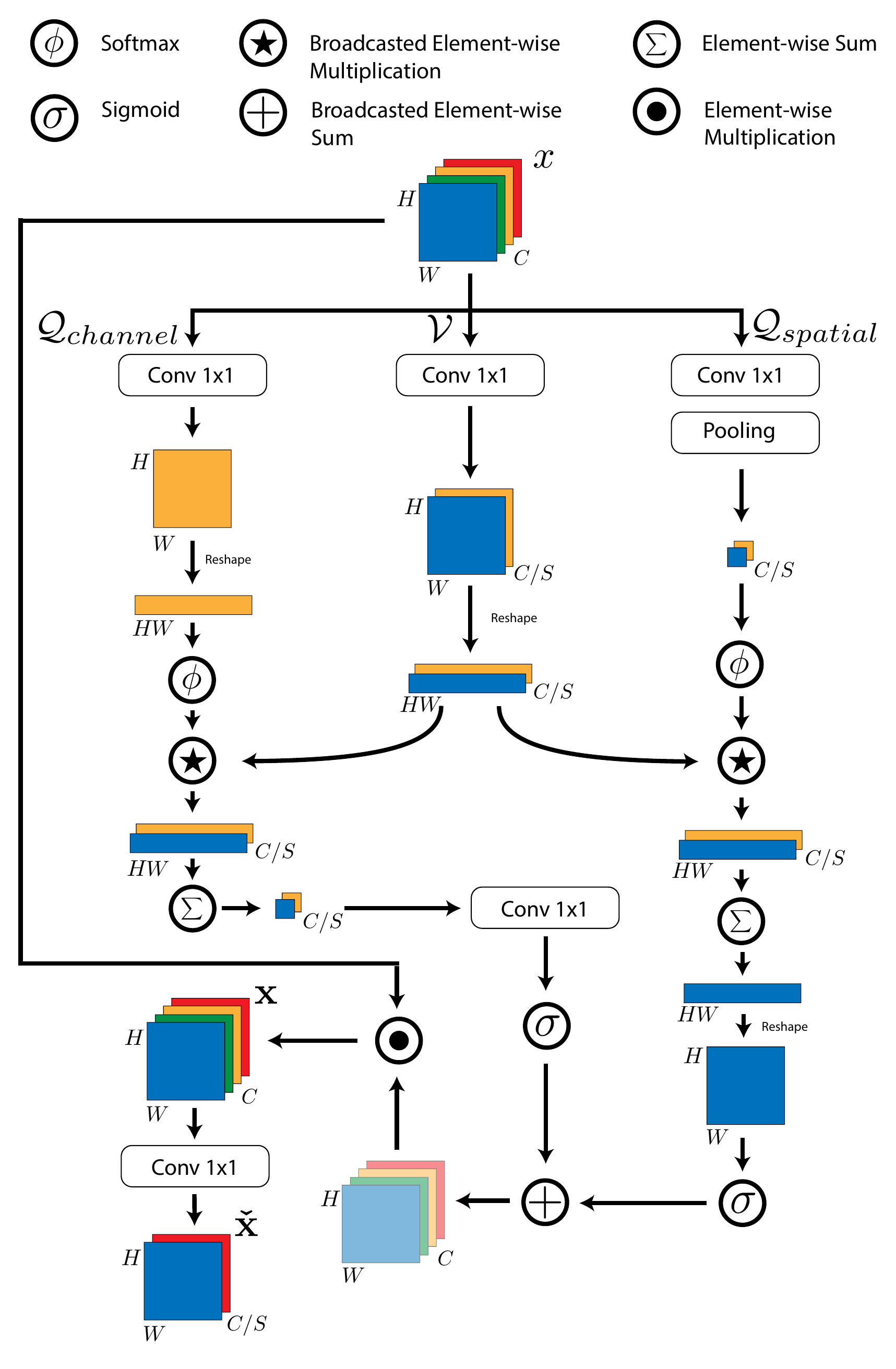}
    \vspace{-4mm}
     \caption{\textbf{Fast Mixed Filtration.} This block serves as a lightweight and effective attention mechanism. The input $x$ is filtered to produce the compressed representations $\bold{\check{x}}$. The broadcast and element-wise operations make this block efficient on CPU. }
     \label{fig:attention2}
     \vspace{-5mm}
\end{figure}
The dual-template features and the dual-search-region features used by the correlation module $CC(\cdot,\cdot)$ could each be the naive concatenation of the respective backbone features in each pair. However, we can potentially improve their combination by processing them with an efficient learnable layer that filters out less useful components while enhancing those important for the task at hand. This should hopefully lead to improved performance. In \Cref{fig:all_ablation}(top), we have shown just that. Especially when tracking in the OOD case, naive feature concatenation severely underperforms the filtered combination.

We consider filtration mechanisms such as self-attention~\cite{vaswani2017attention}, only tailored to convolutional models since they tend to be more efficient on CPU. Polarized Self-Attention (PSA)~\cite{liu2021polarized} stands out as it functions as a filter and is a more powerful variant than CBAM~\cite{woo2018cbam}, with time complexity of $O(CWH)$, where $C$, $W$, $H$, are channel, width, and height of the tensor respectively. Another notable option is the CNN-ViT-based attention block MobileViTv3~\cite{wadekar2022mobilevitv3}, based on separable attention~\cite{mehta2022separable}. As shown in \Cref{tab:att_comp}, we observed limited performance gain of MobileViTv3 over PSA, with relatively high FLOPs, parameters, and latency. We also noticed that PSA performs several matrix multiplications causing the latency on CPU to still be considerably high. This motivated the development of our \emph{Fast Mixed Filtration (FMF)}, a new and more efficient mixed filtration method. It is defined as follows
\begin{equation}\small
    \centering
    \label{eq:fast_pol_attention}
        \begin{aligned}   
            A_{ch}(x) &= \sigma(W_{ch}(\sum^{HW} W^V(x) \star \phi(W^Q_{ch}(x))  )) \; , \\ 
            A_{sp}(x) &= \sigma(\sum^{C} \phi(W^Q_{sp}(x)) \star W^V(x))\; ,\\ 
            \bold{x} &= (A_{ch} \oplus  A_{sp}) \odot x \; , \\
\end{aligned}
\end{equation}
where the notation means the following: $\sigma :=$ sigmoid, $\phi :=$ softmax, $W := \mathrm{conv}1 \times 1$,  $\star :=$ broadcasted element-wise multiplication, $\oplus :=$ broadcasted element-wise sum, $\odot :=$ element-wise multiplication, $\sum :=$ element-wise sum, $A_{ch} :=$ channel filter, $A_{sp} :=$ spatial filter, $V:=$ values, $Q:=$ queries, $x :=$ input  and $\bold{x} :=$ output. \Cref{fig:attention2} depicts the schematic computation of the FMF layer.

Differently than PSA, the improved efficiency of FMF stems from reducing the computation overhead in \Cref{eq:fast_pol_attention} by limiting the matrix operations to be broadcasted element-wise multiplications and element-wise summations across the vectors, and by setting $W^V_{sp}=W^V_{ch}=W^V$ for both $A_{ch}$ and $A_{sp}$, which further reduces the number of parameters. Similar to CBAM and PSA, we utilize a squeeze-and-excite framework~\cite{iandola2016squeezenet} to excite the relevant features within this block. As shown in \Cref{tab:att_comp}, the latency of FMF is 0.4ms on a CPU, which has decreased from the 0.8ms of PSA, while not experiencing any performance loss.

We use FMF with a squeeze rate $S=2$ (see \Cref{fig:attention2}). In this way, if $x$ is the concatenated representation of the dual-template (i.e., $x_T \doteq F_D \cup F_T$) or of the dual-search-region (i.e., $x_t \doteq F_t \cup F_S$), then $\bold{x}$ is the filtered representation with channel dimension $2C$, to which we apply a channel dimension reduction from $2C$ to $C$, giving us $\bold{\check{x}}$ (i.e., $\bold{\check{x}}_T$ or $\bold{\check{x}}_t$).

\textsl{\textbf{Pixel-wise Cross-Correlation.}}
We combine the filtered dual representations $\bold{\check{x}}_T$ and $\bold{\check{x}}_t$ with the block $CC(\cdot)$, which is a pixel-wise cross-correlation. The output is then concatenated with $F_t$ and fed to a one-layer convolution to reduce the number of channels processed by the heads block.

\textbf{Heads Block.}
Similar to~\cite{zhang2020ocean, borsuk2022fear}, tracking output is given by a classification head $CH(\cdot)$, and a bounding box regression head $BH(\cdot)$.
We use 2 lightweight convolution layers for $CH(\cdot)$ and 4 for $BH(\cdot)$.
The last layer of $CH(\cdot)$ has only 1 channel and predicts the foreground/background confidence score for the object, whereas the last layer of $BH(\cdot)$ has 4 channels, each responsible for predicting separate values ($x_{min}$, $y_{min}$, $x_{max}$, and $y_{max}$ for the object bounding box in the frame at time $t$), which is why we choose a higher number of convolution blocks for $BH(\cdot)$. Following~\cite{borsuk2022fear}, we keep the spatial resolution of these maps to 16 $\times$ 16. 

\begin{table*}[t]
  \caption{Comparison of FLOPs, number of parameters, and latency when using MobileViTv3\cite{wadekar2022mobilevitv3}, Polarized Self-Attention\cite{liu2021polarized} and Fast Mixed Filtration, with their performances on AVisT \cite{noman2022avist} and LaSOT \cite{fan2021lasot}.}
  \vspace{-3mm}
    \label{tab:att_comp}
    \centering
    \scalebox{.65}{
    \begin{tabular}{c|c |c |c c c |c c | c c}
    \hline
    \multirow{2}{*}{Attention} & \multirow{2}{*}{FLOPs (G.)($\downarrow$)} & \multirow{2}{*}{Params (M.)($\downarrow$)} & \multicolumn{3}{c|}{Latency (ms)($\downarrow$)} & \multicolumn{2}{c|}{AVisT}   & \multicolumn{2}{c}{LaSOT} \\
    & & & CPU & GPU & Nano & AUC($\uparrow$) & OP50($\uparrow$) & AUC($\uparrow$)& Prec.($\uparrow$)\\
    \hline
    MobileViTv3 \cite{wadekar2022mobilevitv3} & 0.220 & 0.862 & 2.7 & 1.3 & 6.5 & 0.435 & 0.490 & 0.568 & 0.588 \\
    PSA \cite{liu2021polarized} & 0.051 & 0.526 & 0.8 & 0.25 & 4.3  & 0.457 & 0.529  & 0.567 & 0.588 \\
    \rowcolor{lightgray!20}FMF (ours) & 0.034 & 0.395 & 0.4 & 0.2 & 3.5 & 0.458 & 0.529 & 0.572 & 0.592 \\
    \hline
    \end{tabular}
    }
    \vspace{-5mm}
\end{table*}

\subsection{Training Losses}

\textbf{Transitive Relation Loss (TRL).}
We help focussing the filtered representations of the dual-template and the dual-search-region on providing relational information that will aid the downstream tasks. To this end, we recognize that $\Omega(F_D,F_T)$ and $\Omega(F_t,F_S)$ should be ``similar'', since all the inputs contain information about the object. We introduce a loss  $\mathcal{L}_{TR}$ to encourage that. At the same time, given the high built in similarity between $F_D$ and $F_S$ (since $I_D \subset I_S$), to avoid learning representations that ignore the static template ($F_T$) and search region ($F_t$) information, we also add a regularization loss $\mathcal{L}_{Reg}$ that pulls the representations close to $F_t$. $\mathcal{L}_{Reg}$ is applied between $\Omega(F_D,F_T)$ and $F_t$.
$\mathcal{L}_{TR}$ and $\mathcal{L}_{Reg}$ are defined as
\begin{equation}\small
    \centering
    \label{eq:relation}
        \begin{aligned} 
            \mathcal{L}_{TR}&=\mathcal{D}(\Omega(F_D,F_T),\Omega(F_t,F_S)) \; , \\
            \mathcal{L}_{Reg}&=\mathcal{D}(\Omega(F_D,F_T),F_t)\; , \\
            \mathcal{D}(x_1, x_2)&=\dfrac{1}{2} (D(h_1(x_1), h_2(x_2)) +  D(h_1(x_2), h_2(x_1))), \\
            D(z_1,z_2)&=1 - \dfrac{z_1}{{||z_1||}_2} \cdot \dfrac{z_2}{{||z_2||}_2} \; ,
        \end{aligned}
\end{equation}
where $||\cdot||_2$ is the $\ell_2$-norm,  $h_1$ and $h_2$ are MLP projection heads used only during training, and $D(\cdot, \cdot)$ calculates the cosine distance. Moreover, $\mathcal{D}(\cdot,\cdot)$ is computed as in~\cite{chen2021exploring}, where we implement the stop-gradient operation to avoid degenerated solutions. We refer to the pair $\mathcal{L}_{TR}$ and $\mathcal{L}_{Reg}$ as the \emph{transitive relation loss (TRL)} since it is meant to bridge the similarities between template and search region images with the aid of the dynamic components so that the relevant relational differences can be highlighted by the downstream blocks for task purposes. The TRL loss improves tracking performance, especially when both components are used. See \Cref{fig:all_ablation}(middle). Notably, in the OOD case, the AVisT \cite{noman2022avist} AUC improves from 43.7\% to 45.8\%.

\textbf{Total Tracking Loss.} We use standard losses for the regression and classification heads. We use the $IoU$ loss~\cite{rezatofighi2019generalized}, $\mathcal{L}_{IoU}$, for the bounding box regression, $BH(\cdot)$, and the focal loss~\cite{lin2017focal}, $\mathcal{L}_{FL}$, for the classification head, $CH(\cdot)$. We refer to \cite{rezatofighi2019generalized,lin2017focal} for their definition.
The offline training of the tracker is therefore based on the \emph{total tracking loss}
\begin{equation} \small
    \centering
    \label{eq:overall_loss}
            \mathcal{L} =  \mathcal{L}_{IoU} + \lambda_{FL} \mathcal{L}_{FL} + \lambda_{TR} \mathcal{L}_{TR} + \lambda_{Reg} \mathcal{L}_{Reg},
\end{equation}
where we set $\lambda_{FL}$, $\lambda_{TR}$, and $\lambda_{Reg}$ to 1, 1/3, and 1/3, respectively.

\textbf{Dynamic Update.} 
Different strategies can be implemented for when to update the dynamic image template and dynamic search region image. \Cref{tab:component_ablation_updates} reports our case-study focussing on parameter-free strategies, which suggests that different strategies are specific to their approaches and are not generally applicable. A simple yet effective strategy that gave us reliable performance is based on maintaining a running average of the classification scores $\overline{\rho }_t$
\begin{equation}
    \label{eq:sample_update}
            \overline{\rho }_t = (1-{\lambda}_{D}) \overline{\rho}_{t-1} +  {\lambda}_{D} {\rho}_t \; ,
\end{equation}
where, ${\rho}_t$ is the score at time $t$, and ${\lambda}_{D}$ is a momentum parameter, which we set to 0.25. We also start a counter, $C$, that when it reaches, let us say $N=60$ frames ($\approx$ 2 seconds), we compare the current classification score with the running average, and if ${\rho}_t > \overline{\rho }_{t-1}$, then we update the dynamic image components and reset the counter, otherwise we repeat the test at the next iteration. This strategy is effective, parameterless, and uses limited computational resources.
  
\subsection{Dynamic Test-Time Adaptation} \label{sec:methods_tta}

To increase tracking performance, especially at OOD test-time, we introduce a dynamic test-time adaptation procedure based on a batch normalization (BN) correction tailored specifically to tracking. We applied this strategy to the classification and bounding box regression heads. First, BN layers are generally computed by the following equations:
\begin{minipage}{.5\linewidth}
    \begin{equation}\small
        \centering
        \label{eq:bn}
            \begin{aligned}  
            BN(x) = \gamma \frac{x - E(x)}{\sqrt{Var(x)}} +  \beta ,
            \end{aligned}
    \end{equation}
    \end{minipage}%
    \begin{minipage}{.5\linewidth}
        \begin{equation}\small
            \centering
            \label{eq:bn_running}
                \begin{aligned}  
                    \overline{\mu}_t = (1-\alpha)  \overline{\mu}_{t-1} +  \alpha  {\mu}_t ,\\
                    \overline{\sigma}_{t}^2 = (1-\alpha) \overline{\sigma}_{t-1}^2 +  \alpha {\sigma}_{t}^2 ,
                \end{aligned}
        \end{equation}
\end{minipage}
where $x$ is the input feature, $E(x)$ and $Var(x)$ are the expected value and variance of $x$. $\gamma$ and $\beta$ are learnable parameters for scaling and shifting. The BN layers keep track of the running mean and variance through \Cref{eq:bn_running}, where ${\mu}_t$ and ${\sigma}_{t}$ are current expected value and variance respectively, and $\overline{\mu}_t$ and $\overline{\sigma}_{t}$ are used for $E(x)$ and $Var(x)$, respectively. $\alpha$ is the momentum parameter. 
Estimating learnable parameters $\gamma$ and $\beta$ during testing would require a backward pass, with great detriment to the speed. Therefore, we propose a method that dynamically updates the BN statistics $E(x)$ and $Var(x)$ during testing, which has negligible computational overhead for maintaining speed.
\begin{table*}[t!]
  \caption{Comparative study on VOT2020 Benchmark \cite{kristan2020eighth}. Red, blue, and green colors describe the best three CPU real-time trackers whereas bold suggests the best CPU non-real-time tracker.}
  \vspace{-3mm}
  \label{tab:vot2020_res}
  \centering
  \scalebox{0.6}{
  
  \begin{tabular}{c|c c c |c c c c c  >{\columncolor{lightgray!20}}c}
    \hline
    & \multicolumn{3}{c|}{CPU non-real-time Methods}   & \multicolumn{6}{c}{CPU real-time Methods} \\

    Trackers  & STARK-ST50 \cite{yan2021learning} & STARK-S50 \cite{yan2021learning}  & DiMP \cite{bhat2019learning} &  HCAT \cite{chen2022efficient}  &  E.T.Track \cite{blatter2023efficient}  & LightTrack \cite{yan2021lighttrack} & ATOM \cite{danelljan2019atom} & MixFormerV2-S \cite{cui2024mixformerv2}& \bf{S-Tiny} \\
    \hline
    EAO         & \bf{0.308} & 0.280 & 0.274 & \color{RoyalBlue} \bf{0.276} & 0.267 & 0.242 & \color{ForestGreen} \bf{0.271} & 0.258 & \color{Mahogany} \bf{0.291} \\
    Accuracy    & \bf{0.478} & 0.477 & 0.457 & \color{ForestGreen} \bf{0.455} & 0.432 & 0.422 & \color{RoyalBlue} \bf{0.462} & - & \color{Mahogany} \bf{0.491}\\
    Robustness  & \bf{0.799} & 0.728 & 0.740 & \color{Mahogany} \bf{0.747} & \color{RoyalBlue} \bf{0.741} & 0.689 & \color{ForestGreen} \bf{0.734} & - & \color{RoyalBlue} \bf{0.741}\\
    \hline
    FPS (CPU)   & 6 & 7 & \bf{14} & \color{ForestGreen} \bf{60} & 35 & \color{RoyalBlue} \bf{67} & 30 & 37 & \color{Mahogany} \bf{100} \\
    FPS (GPU)   & 66 & 66 & \bf{127} & \color{ForestGreen} \bf{300} & 108 & 170 & 240 & \color{RoyalBlue} \bf{420} & \color{Mahogany} \bf{425} \\
    FPS (Nano)   & 10 & \bf{12} & 10 & \color{RoyalBlue} \bf{24} & 10 & \color{ForestGreen} \bf{17} & 13 & \color{Mahogany} \bf{40} & \color{Mahogany} \bf{40} \\
    \hline

  \end{tabular}
  }
\end{table*}

Prior works have explored BN adaptation for classification purposes, with some using a backward pass for adaptation \cite{niu2022efficient, wang2020tent}, while others introduced backward-free adaptation \cite{pan2018two, schneider2020improving, mirza2022norm, li2016revisiting}. However, none are directly applicable for tracking. As we show in \Cref{tab:tta_res}, applying a new Instance-Norm (IN) layer \cite{pan2018two} is expensive, and slows down the speed twofold without noticeable improvement. Additionally, the batch size for tracking remains 1, which is too small to make any significant improvement with \cite{schneider2020improving} that applies weighted momentum based on the target batch-size. Given the scale of our architecture and limited number of parameters, by replacing batch statistics with instance statistics, AdaBN \cite{li2016revisiting} does not improve the performance either. DUA \cite{mirza2022norm} uses source statistics as a prior for the incoming task but does not stay anchored to the source statistics, resulting in target statistics drifting away from the original distribution, causing performance drop. Therefore, we propose that BN statistics should be updated with weighted instance statistics while remaining anchored to the source statistics. This results in the following strategy
\begin{equation}
    \centering
    \label{eq:bn_update}
        \begin{aligned}  
            \overline{\mu}_{I,t} &= (1-\lambda{_{BN}}) \overline{\mu} +  \lambda{_{BN}} \mu_{I,t} \; ,\\
            \overline{\sigma}_{I,t}^2 &= (1-\lambda{_{BN}}) \overline{\sigma}^2 +  \lambda{_{BN}} \sigma_{I,t}^2 \; ,
        \end{aligned}
\end{equation}
where $\overline{\mu}$ and $\overline{\sigma}^2$ are the final running mean and variance of the model trained on the source data, respectively, and $\mu_{I,t}$, and $\sigma_{I,t}^2$ are mean and variance calculated from the instance at time $t$, respectively. $\overline{\mu}_{I,t}$ and $\overline{\sigma}_{I,t}^2$ are updated based on the current instance and used for feature normalization at time $t$. $ \lambda{_{BN}}$ is set to 0.1. This backward-free \emph{dynamic test-time adaptation (DTTA)} strategy is efficient with negligible difference in latency as shown in \Cref{tab:tta_res}.


\begin{table*}[th!]
  \caption{Comparative Study with other SOTA approaches on various benchmarks including AVisT\cite{noman2022avist}, NFS30\cite{kiani2017need}, UAV123\cite{mueller2016benchmark}, TrackingNet\cite{muller2018trackingnet}, GOT-10k\cite{Huang2021}, and LaSOT\cite{fan2021lasot}. Red, blue, and green colors describe the best three CPU real-time trackers whereas bold suggests the best CPU non-real-time tracker.
  }
  \vspace{-3mm}
  \label{tab:main_res}
  \centering
  \scalebox{0.65}{
    \setlength{\tabcolsep}{0.45em}
  \begin{tabular}{c |c c  c  |c c  |c c |c c  c |c c |c c |c c c}
  
  \hline
  & \multicolumn{7}{c|}{Out-of-Distribution (OOD) test sets} & \multicolumn{7}{c|}{In-Distribution (ID) test sets} & & \\
  \multirow{2}{*}{Methods}  & \multicolumn{3}{c|}{AVisT\cite{noman2022avist}} & \multicolumn{2}{c|}{NFS30\cite{kiani2017need}} & \multicolumn{2}{c|}{UAV123\cite{mueller2016benchmark}} & \multicolumn{3}{c|}{TrackingNet\cite{muller2018trackingnet}} & \multicolumn{2}{c|}{GOT-10k\cite{Huang2021}} & \multicolumn{2}{c|}{LaSOT\cite{fan2021lasot}}  & \multicolumn{3}{c}{ FPS ($\uparrow$)}  \\
   & AUC & OP50 & OP75    & AUC & Prec. & AUC & Prec. & AUC & P$_{norm}$ & Prec. & AO & SR$_{0.50}$ & AUC & Prec. & CPU & GPU & Nano\\
   
  \hline
  \multicolumn{18}{c}{CPU non-real-time Methods} \\
  \hline

  STARK-ST50 \cite{yan2021learning}         & 0.511 & 0.592 & \bf{0.391}       & 0.652  & - & 0.691  & - & 0.813  & 0.861  & - & 0.680  & 0.777 & 0.666  & -  & \bf{7} & 66 & 10\\
  Ocean \cite{zhang2020ocean}               & 0.389 & 0.436 & 0.205       & 0.573  & 0.706 & 0.574  & - & -  & -  & - & 0.611  & 0.634  & 0.505  & 0.517 & 2  & 70 & 18\\
  SiamRPN++ \cite{li2019siamrpn++}          & 0.390 & 0.435 & 0.212       & 0.596  & \bf{0.720} & 0.593  & - & 0.733  & 0.800  & - & -  &  - & 0.503  & 0.496  & 1.4 & 145 & 10 \\
  SiamMask \cite{wang2019fast}              & 0.358 & 0.401 & 0.185       & - & -  & - & -  & -  & - & -  & -  & - & - & -  & 4 & \bf{308} & 20 \\
  SiamBAN \cite{chen2020siamese}            & 0.376 & 0.432 & 0.217       & 0.594  & - & 0.631  & 0.833  & -  & -  & - & -  & - & 0.514  & 0.598 & 4 & 300 & \bf{24} \\
  SeqTrack-L384 \cite{chen2023seqtrack}     & - & - & -                   & 0.662  & - & 0.685  & - & \bf{0.855}  & \bf{0.895}  & \bf{0.858} & 0.748  & 0.819  & \bf{0.725}  & 0.793  & 0.4 & 15 & -\\
  MixFormerV2-B \cite{cui2024mixformerv2}   & - & - & -                   & -  & - & \bf{0.699}  & \bf{0.921} & 0.834  & 0.881  & 0.816 & 0.739  & -  & 0.706  & \bf{0.808}  & \bf{7} & 130 & 15 \\
  DropMAE \cite{wu2023dropmae}              & - & - & -                   & -  & - & -  & - & 0.841  & 0.889  & - & \bf{0.759}  & \bf{0.868}  & 0.718  & 0.780  & 4 & 98 & 10 \\
  TransT \cite{chen2021transformer}         & 0.490 & 0.564 & 0.372       & 0.657  & - & 0.691  & - & 0.814  & 0.867  & 0.803 & 0.723  & 0.824   & 0.649  & 0.690  & \bf{7} & 85 & 8 \\
  OSTrack-256 \cite{ye2022joint}            & - & - & -                   & 0.647  & - & 0.683  & - & 0.831  & 0.878  & 0.820 & 0.710  & 0.804  & 0.691  & 0.752 & 4 & 98 & 18 \\
  ToMP-50 \cite{mayer2022transforming}      & \bf{0.516} & \bf{0.595} & 0.389      & \bf{0.669}  & - & 0.690  & - & 0.786  & 0.862  & 0.812 & -  & - & 0.676  & 0.722  & \bf{7} & 83 & 6 \\
  
  \hline
  \multicolumn{18}{c}{CPU real-time Methods} \\
  \hline  
  
  HiT-Small~\cite{kang2023exploring} 
   & - & - & - &  0.618 & - & 0.633 & - & \color{ForestGreen} \bf{0.777} & 0.819 & \color{ForestGreen} \bf{0.731} & 0.626 & 0.712 & 0.605 & \color{ForestGreen} \bf{0.615} & - & - & -\\
  SMAT \cite{gopal2024separable} & \color{ForestGreen} \bf{0.447} & \color{ForestGreen} \bf{0.507} & \color{ForestGreen} \bf{0.313}  &  \color{RoyalBlue} \bf{0.620}  &  \color{RoyalBlue} \bf{0.746} & \color{ForestGreen} \bf{0.643}  & \color{ForestGreen} \bf{0.839}  &  \color{Mahogany} \bf{0.786} &  \color{Mahogany} \bf{0.842}  &  \color{Mahogany} \bf{0.756} &  \color{RoyalBlue} \bf{0.645}  &  \color{RoyalBlue} \bf{0.747} &  \color{Mahogany} \bf{0.617}  &  \color{Mahogany} \bf{0.646}  & 34 & 158 & 20 \\
  E.T.Track \cite{blatter2023efficient} & 0.390 & 0.412 & 0.227               & 0.570  & 0.694 & 0.623  & 0.806 & 0.745  & 0.798  & 0.698 & 0.566  & 0.646 & 0.589  & 0.603 & 35 & 108 & 10 \\
  MixFormerV2-S \cite{cui2024mixformerv2} & 0.396 & 0.425 & 0.227             & 0.610  & 0.722 & 0.634  & 0.837 & 0.758  & 0.811  & 0.704 & 0.587  & 0.672 & \color{ForestGreen} \bf{0.606}  & 0.604  & 37 & \color{ForestGreen} \bf{420} & \color{Mahogany} \bf{40} \\
  HCAT \cite{chen2022efficient} & 0.418 & 0.481 & 0.263                       & \color{ForestGreen} \bf{0.619}  & 0.741 & 0.636  & 0.805 & 0.766  & \color{ForestGreen} \bf{0.826}  & 0.729 & \color{ForestGreen} \bf{0.634}  & \color{ForestGreen} \bf{0.743} & 0.590  & 0.605 & \color{ForestGreen} \bf{60} & 300 & \color{ForestGreen} \bf{24} \\
  LightTrack \cite{yan2021lighttrack} & 0.404 & 0.437 & 0.242                 & 0.565  & 0.692 & 0.617  & 0.799 & 0.729  & 0.793  & 0.699 & 0.582  & 0.660  & 0.522  & 0.517 & \color{RoyalBlue} \bf{67} & 170 & 17 \\
  FEAR-XS \cite{borsuk2022fear} & 0.387 & 0.421 & 0.220                       & 0.486  & 0.563 & 0.610  & 0.816 & 0.715  & 0.805 & 0.699 & 0.573  & 0.681  & 0.535  & 0.545 &  \color{Mahogany} \bf{100} & \color{Mahogany} \bf{450} & \color{Mahogany} \bf{40}\\

  \rowcolor{lightgray!20} \bf{S-Tiny}  &  \color{RoyalBlue} \bf{0.472} &  \color{RoyalBlue} \bf{0.543} &  \color{RoyalBlue} \bf{0.353}  &  \color{RoyalBlue} \bf{0.620}  &  \color{Mahogany} \bf{0.747} &  \color{RoyalBlue} \bf{0.662} &  \color{RoyalBlue} \bf{0.856} & 0.741  & 0.819  & 0.720 & 0.614  & 0.728 & 0.590  &  0.607  &  \color{Mahogany} \bf{100} & \color{RoyalBlue} \bf{425} & \color{Mahogany} \bf{40}\\
  \rowcolor{lightgray!20} \bf{S-Small} &  \color{Mahogany} \bf{0.479} &  \color{Mahogany} \bf{0.557} &  \color{Mahogany} \bf{0.372}  &  \color{Mahogany} \bf{0.624}  & \color{ForestGreen} \bf{0.744} &  \color{Mahogany} \bf{0.681}  &  \color{Mahogany} \bf{0.858} & \color{RoyalBlue} \bf{0.784}  & \color{RoyalBlue} \bf{0.835}  & \color{RoyalBlue} \bf{0.746} & \color{Mahogany} \bf{0.646}  &  \color{Mahogany} \bf{0.751}  & \color{RoyalBlue} \bf{0.607}  & \color{RoyalBlue} \bf{0.622}  &  45 & 400 & \color{RoyalBlue} \bf{30}\\
  \hline
  \end{tabular}
  }
  \vspace{-4mm}
\end{table*}

\begin{table*}[h]
  \caption{Comparative study on ITB \cite{li2021informative}, OTB \cite{7001050}, TC128 \cite{liang2015encoding}, and  DTB70 \cite{drone-tracking} benchmarks in terms of their AUC score. Red, blue, and green colors describe the best three CPU real-time trackers whereas bold suggests the best CPU non-real-time tracker.}
  \vspace{-3mm}
  \label{tab:itb_otb_tc_dtb_res}
  \centering
  \scalebox{0.65}{
  
  \begin{tabular}{c|c c c c c c |c c c >{\columncolor{lightgray!20}}c >{\columncolor{lightgray!20}}c}
    \hline
    & \multicolumn{6}{c|}{CPU non-real-time Methods}   & \multicolumn{5}{c}{CPU real-time Methods} \\

     & DropMAE \cite{wu2023dropmae}  & TransT  \cite{chen2021transformer} & STARK \cite{yan2021learning}  & DiMP \cite{bhat2019learning}  & SiamRPN++ \cite{li2019siamrpn++} & Ocean \cite{zhang2020ocean} &   E.T.Track \cite{blatter2023efficient}  & LightTrack \cite{yan2021lighttrack} & ATOM \cite{danelljan2019atom} & \bf{S-Tiny} & \bf{S-Small}  \\
    \hline
    ITB \cite{li2021informative} & \bf{0.650} & 0.547 & 0.576 & 0.537 & 0.441 & 0.477 & - & - & \color{ForestGreen} \bf{0.472} & \color{RoyalBlue} \bf{0.548} & \color{Mahogany} \bf{0.555} \\
    OTB \cite{7001050} & \bf{0.696} & 0.695 & 0.681 & 0.684 & 0.687 & 0.684 & \color{ForestGreen} \bf{0.678} & 0.662 & 0.669 & \color{RoyalBlue} \bf{0.709} & \color{Mahogany} \bf{0.713} \\
    TC128 \cite{liang2015encoding} & - & 0.596 & \bf{0.626} & 0.612 & 0.577 & 0.557 & - & 0.550 & \color{ForestGreen} \bf{0.599} & \color{RoyalBlue} \bf{0.617} & \color{Mahogany} \bf{0.630} \\
    DTB70 \cite{drone-tracking} & - & \bf{0.667} & 0.638 & - & 0.569 & 0.455 & - & \color{ForestGreen} \bf{0.491} & - & \color{RoyalBlue} \bf{0.656} & \color{Mahogany} \bf{0.662} \\
    \hline
    FPS (CPU) & 4 & 8 & 7 & \bf{14} & 1.4 & 2 & 35 & \color{RoyalBlue} \bf{67} & 30 & \color{Mahogany} \bf{100} & \color{ForestGreen} \bf{45} \\
    FPS (GPU) & 85 & 66 & 66 & 127 & \bf{145} & 70 & 108 & 170 & \color{ForestGreen} \bf{240} & \color{Mahogany} \bf{425} & \color{RoyalBlue} \bf{400} \\
    FPS (Nano) & 10 & 8 & 12 & 10 & 10 & \bf{18} & 10 & \color{ForestGreen} \bf{17} & 13 & \color{Mahogany} \bf{40} & \color{RoyalBlue} \bf{30} \\
    \hline

  \end{tabular}
}
\end{table*}

  \begin{table*}[h]
    \caption{Comparative study on test-time adaptation (TTA) approaches on AVisT\cite{noman2022avist} as it involves various extreme distribution shifts with real-world corruptions and ITB \cite{li2021informative} as the next most challenging benchmark. The best results are in bold.
    }
    \label{tab:tta_res}
    \vspace{-3mm}
    \centering
    \scalebox{0.65}{
    \begin{tabular}{c|c c  c c |c c  c c |c c c}
    
    \hline
    
    \multirow{2}{*}{Methods}  & \multicolumn{4}{c|}{AVisT\cite{noman2022avist}}  & \multicolumn{4}{c|}{ITB\cite{li2021informative}} &  \multicolumn{3}{c}{Latency (ms) $\downarrow$ }  \\
     & AUC & OP50 & OP75 & Prec.    & AUC & OP50 & OP75 & Prec. & CPU & GPU & Nano\\
  
    \hline 
    No TTA & 0.458 & 0.529 & 0.340 & 0.413 & 0.539 & 0.659 & 0.483 & 0.631 & \bf{3.6} & \bf{0.6} & \bf{9.8}\\
    \hline
    \multicolumn{12}{c}{Backward-Based Methods} \\
    \hline
    TENT \cite{wang2020tent} & 0.460 & 0.518 & 0.356 & 0.417 & 0.530 & 0.635 & 0.472 & 0.610 & 9.9 & 2.9 & 30.4 \\
    ETA \cite{niu2022efficient} & 0.459 & 0.516 & \bf{0.360} & 0.416 & 0.525 & 0.630 & 0.470 & 0.606 & 9.9 & 2.8 & 35.6 \\
    \hline
    \multicolumn{12}{c}{Backward-Free Methods} \\
    \hline
    Momentum \cite{schneider2020improving} & 0.452 & 0.513 & 0.341 & 0.411 & 0.540 & 0.656 & 0.484 & 0.632 & 3.7 & 0.7 & 15.1 \\
    DUA \cite{mirza2022norm} & 0.427 & 0.484 & 0.307 & 0.394 & 0.516 & 0.628 & 0.467 & 0.591 & 3.7 & 0.7 & 15.1 \\
    IN \cite{pan2018two} & 0.454 & 0.515 & 0.342 & 0.417 & 0.519 & 0.628 & 0.455 & 0.604 & 6.6 & 1.9 & 23.1 \\
    AdaBN \cite{li2016revisiting} & 0.456 & 0.517 & 0.345 & 0.414 & 0.522 & 0.632 & 0.461 & 0.608 & 3.7 & 0.7 & 15.1 \\
    \rowcolor{lightgray!20} DTTA (ours) & \bf{0.472} & \bf{0.543} & 0.353 & \bf{0.440} & \bf{0.548} & \bf{0.667} & \bf{0.494} & \bf{0.634} & 3.7 & 0.7 & 15.1 \\
    \hline
    \end{tabular}
    }
    \vspace{-5mm}
  \end{table*}

\section{Experiments}

\label{sec:implementation}

\textbf{Model Details.}
With the Tiny backbone, SiamABC consists of 2.03M parameters and uses 0.628 GigaFLOPs. We refer to this tracker as SiamABC-Tiny or S-Tiny. With the Small backbone, SiamABC consists of 9.82M parameters and uses 6.81 GigaFLOPs. We refer to it as SiamABC-Small or S-Small. S-Tiny runs at 100 FPS on a CPU, 425 FPS on a GPU, and 40 FPS on our edge device Jetson Orin Nano, while S-Small runs at 45 FPS (CPU), 400 FPS (GPU), and 30 FPS (Nano).

\textbf{Training.}
All the code is written in PyTorch \cite{paszke2019pytorch}. Both models were trained on a single Nvidia RTX A6000 GPU for 20 epochs. We use a batch size of 32 and ADAM optimizer \cite{kingma2014adam} with a learning rate of $10^{-4}$. We allow close to $10^6$ samples every epoch by randomly sampling sequences and then images from GOT-10k \cite{Huang2021}, LaSOT \cite{fan2021lasot}, COCO2017 \cite{lin2014microsoft}, and TrackingNet \cite{muller2018trackingnet}. We randomly sample a template from a sampled sequence. Further, we randomly choose a search sample from the same sequence with an offset of $\Delta$. The dynamic frames are sampled from the interval between the template and the search samples. We set $\Delta=150$ arbitrarily to facilitate dynamic updates at longer intervals during testing. The input size of the template is $128\times128$, and $256\times256$ for the search frames. For standard augmentations, we crop a template with a size increase offset of 0.2 and a search region with an offset of 2.0. We also apply to the search region crops a random scale and shift factor by uniformly drawing samples from (0.65, 1.35) and (0.92,1.08), respectively. We also apply the color augmentation and use the same post-processing as in~\cite{bertinetto2016fully}.

\textbf{Inference.}
We evaluate all the trackers on two hardware platforms. One is based on an Nvidia RTX 3090 GPU and 12th Gen Intel i9-12900F CPU. The other is an entry-level GPU-based edge device, Nvidia Jetson Orin Nano, which here we abbreviate to `Nano'. All the FPS numbers were reproduced using these two platforms. 

\subsection{Comparison with other Trackers} \label{sec:benchmarks}
We evaluate our SiamABC trackers on 11 challenging benchmarks: AVisT\cite{noman2022avist}, VOT2020\cite{kristan2020eighth}, LaSOT\cite{fan2021lasot}, TrackingNet\cite{muller2018trackingnet}, GOT-10k\cite{Huang2021}, OTB-2015\cite{7001050}, TC128\cite{liang2015encoding}, UAV123\cite{mueller2016benchmark}, NFS30\cite{kiani2017need}, ITB\cite{li2021informative}, and DTB70\cite{drone-tracking}.

\textbf{VOT2020}\cite{kristan2020eighth}  contains 60 challenging videos and employs EAO (expected average overlap) as its metric alongside the accuracy and robustness. In \Cref{tab:vot2020_res}, our tracker, S-Tiny, shows remarkable resilience against difficult scenarios in this benchmark, outperforming HCAT\cite{chen2022efficient} (best real-time method) and STARK-S50\cite{yan2021learning} by 1.05\% and 1.04\% EAO respectively while being more than 14x faster than STARK-S50 on a CPU.

\textbf{AVisT}\cite{noman2022avist} consists of 120 extremely challenging real-world sequences in adverse visibility.
In addition to simple occlusion and fast motion, it involves objects under heavy rain, heavy snow, dense fog, sandstorms, hurricanes, etc. In \Cref{tab:main_res}, we note that this out-of-distribution benchmark highlights a significant performance degradation of SOTA trackers compared to the widely used in-distribution test sets. On the other hand, our trackers, S-Tiny and S-Small, are outperforming HCAT by 5.4\% and 6.1\% respectively and MixFormerV2-S\cite{cui2024mixformerv2} by 7.6\% and 8.3\% respectively, showing our approach's improved ability to track in sequences with adverse conditions. Additionally, S-Tiny outperforms SMAT\cite{gopal2024separable} by 2.5\% while being almost 3x faster on a CPU.

\textbf{UAV123}\cite{mueller2016benchmark} is a benchmark for tracking from an aerial viewpoint involving 123 long video sequences. We outperform HCAT by 2.6\% and 4.5\% respectively, and SMAT by 1.9\% and 3.8\% respectively, confirming the OOD generalization ability of our approach.
\textbf{NFS30}\cite{kiani2017need} is a benchmark collected with extremely high frame rate of 240 FPS for fast tracking. Similar to other approaches, we use the 30 FPS version of the benchmark for evaluation. Our trackers outperform others also in this OOD test. Please refer to \Cref{tab:main_res}.

The \textbf{LaSOT}\cite{fan2021lasot} benchmark involves 280 long test sequences with 2500 frames per sequence on average.
\textbf{TrackingNet}\cite{muller2018trackingnet} is a large benchmark consisting of real-life videos collected from YouTube. There are 511 test videos averging in about 441 frames per sequence. One has to submit the raw data to their evaluation server to obtain the results for a fair evaluation. Similarly, 
\textbf{GOT-10k}\cite{Huang2021} is a challenging short term benchmark consisting of 180 test sequences which are evaluated on their server. Most approaches are trained on the large training sets of LaSOT, TrackingNet, and GOT-10k; thus, the test distributions of such benchmarks remain quite similar. 
Nevertheless, S-Small outperforms most SOTA CPU real-time trackers in these benchmarks while remaining efficient as shown in \Cref{tab:main_res}. Our approach slightly underperforms SMAT
in two ID sets, LaSOT and TrackingNet; however, we note that the trade-off with speed is significant as SMAT runs at 34 FPS (CPU), 158 FPS (GPU), and 20 FPS (Nano), while S-tiny runs at 100 FPS (CPU), 425 FPS (GPU), and 40 FPS (Nano), and S-Small runs at 45 FPS (CPU), 400 FPS (GPU), and 30 FPS (Nano).

As shown in \Cref{tab:itb_otb_tc_dtb_res}, we outperform other CPU-based trackers on the \textbf{OTB-2015}\cite{7001050} benchmark with 100 sequences, even surpassing the CPU non-real-time approaches. 
Another similar benchmark is \textbf{TC128}\cite{liang2015encoding} with 128 challenging color sequences, where we outperform other SOTA approaches as well. \textbf{DTB70}\cite{drone-tracking} is another small-scale UAV benchmark involving 70 long sequences. We consistently show improvement here as well. \textbf{ITB}\cite{li2021informative} is a benchmark with 180 various challenging sequences from many other benchmarks giving an informative evaluation of trackers. ITB is second to AVisT in terms of challenging sequences, and \Cref{tab:itb_otb_tc_dtb_res} confirms a remarkable generalization ability of our approach. We note that DropMAE \cite{wu2023dropmae} was pre-tained on additional diverse training data.

\subsection{Comparision with Adaptation Approaches}\label{sec:tta_res}

In our S-Tiny model we incorporate TENT\cite{wang2020tent} and ETA\cite{niu2022efficient} as our TTA baselines with backward-passes,
where we use our classification output as self-entropy. Additionally, we also evaluated Momentum\cite{schneider2020improving}, DUA\cite{mirza2022norm}, IN\cite{pan2018two}, and AdaBN\cite{li2016revisiting} as our backward-free TTA baselines. We set the batch size to 1. The comparison is shown in \Cref{tab:tta_res} with two of the most challenging benchmarks, AVisT as it involves multiple adverse scenarios with natural corruptions and ITB as the next most challenging benchmark.
TENT and ETA have almost 3x the latency because of the backward passes, and do improve with AVisT. However, we do not observe the same with ITB. Momentum shows negligible improvement, whereas IN, AdaBN, and DUA show performance degradation in both scenarios. When DTTA, our efficient adaptation strategy, is turned on, we notice significant improvement with both benchmarks while having minimal latency.
\begin{table}[t]
  \caption{Case study on parameter-free dynamic updates. }
  \vspace{-3mm}
  \label{tab:component_ablation_updates}
  \centering
  \scalebox{0.65}{
  \begin{tabular}{c|c c | c c | c c c}
  \hline
  \multirow{2}{*}{{\shortstack{ Dynamic Updates \\ (No DTTA)} } }  & \multicolumn{2}{c|}{AVisT}   & \multicolumn{2}{c|}{LaSOT} & \multicolumn{3}{c}{FPS($\uparrow$)} \\
    & AUC & OP50 & AUC & Prec.  & CPU & GPU & Nano \\
  
  \hline
  No Updates & 0.448 & 0.511  & 0.556 & 0.572 & 102 & 430 & 40 \\
  Fixed interval ($N=60$) & 0.445 & 0.514  & 0.519 & 0.533 & 100 & 425 & 40 \\
  FEAR-based \cite{borsuk2022fear} & 0.454 & 0.517  & 0.524 & 0.234 & 96 & 405 & 38 \\
  TATrack-based \cite{he2023target} & 0.361 & 0.377  & 0.414 & 0.430 & 60 & 250 & 22 \\
  \rowcolor{lightgray!20} Ours  & 0.458 & 0.529 & 0.572 & 0.592 & 100 & 425 & 40\\
  \hline
  \end{tabular}
  }
  \vspace{-5mm}
\end{table}

\subsection{Ablation Study}\label{sec:ablation}

\begin{figure}[t]
  \centering
   \includegraphics[width=4cm]{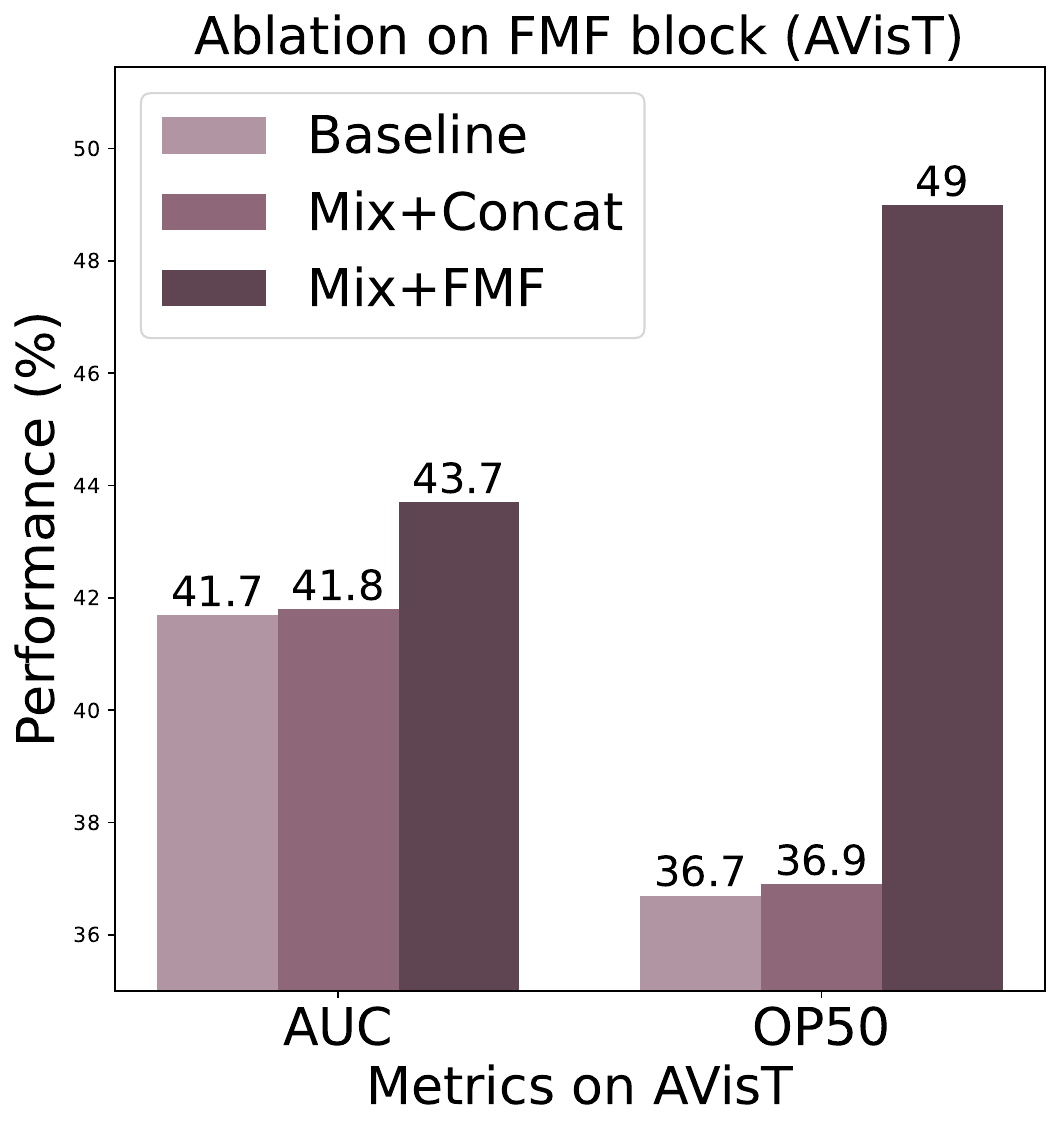}
   \includegraphics[width=4cm]{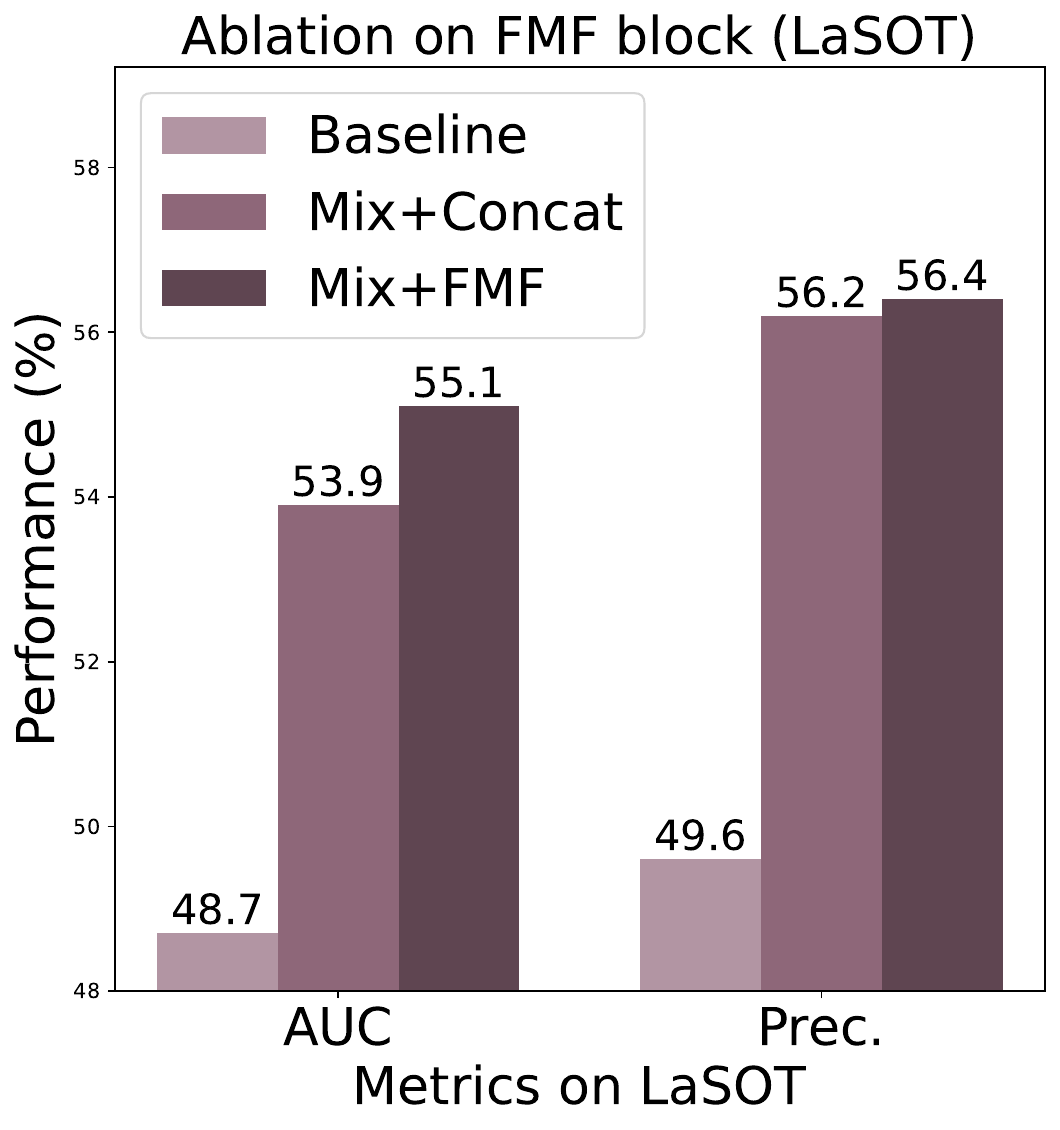}
   \includegraphics[width=4cm]{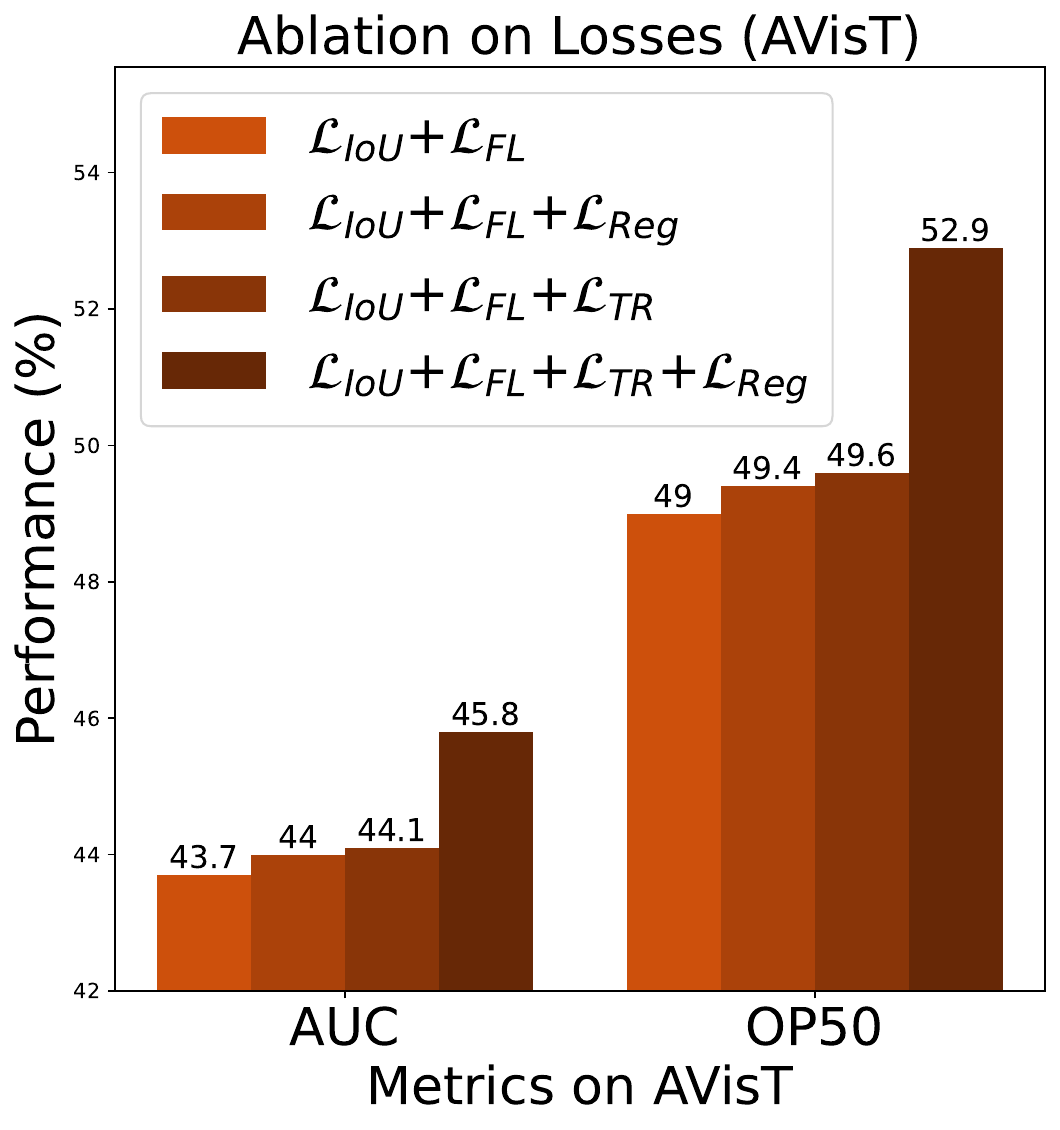}
   \includegraphics[width=4cm]{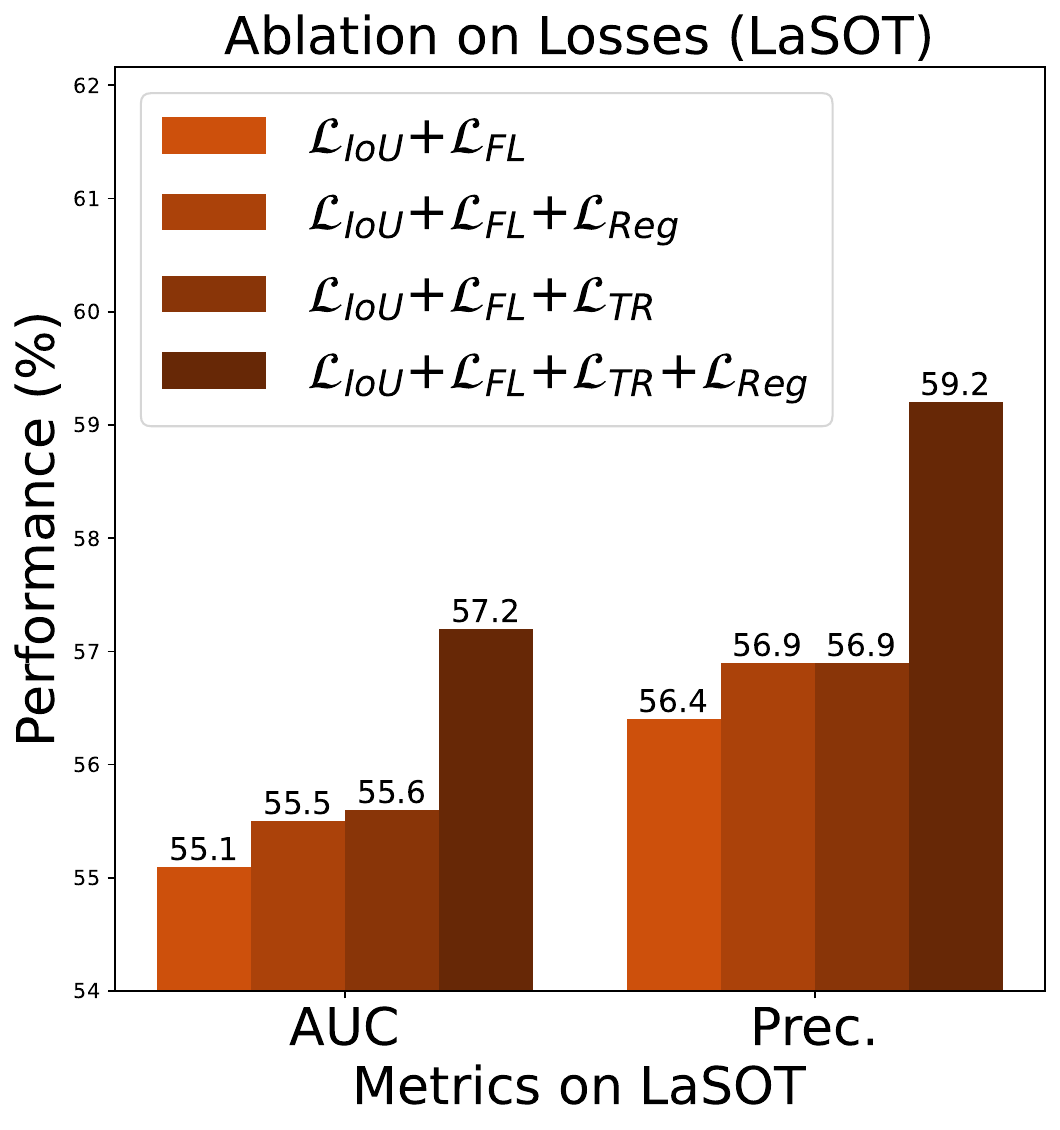}
   \includegraphics[width=4cm]{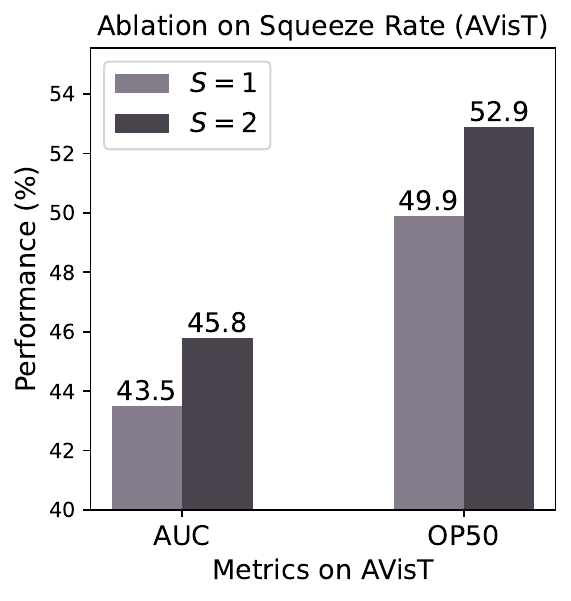}
   \includegraphics[width=4cm]{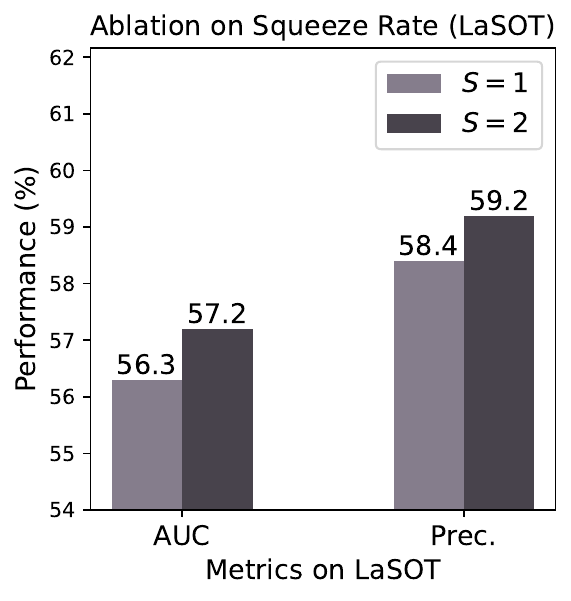}
   \vspace{-4mm}
   \caption{Ablation study on the components of SiamABC-Tiny. Top-row: Ablation on the FMF block. Middle-row: Ablation on TRL losses. Bottom-row: Ablation on squeeze rate.}
   \label{fig:all_ablation}
   \vspace{-5mm}
 \end{figure}

We perform ablation studies on each of the components of S-Tiny used on AVisT as an OOD benchmark and on LaSOT as an ID benchmark. In \Cref{fig:all_ablation}(top), the baseline is obtained by removing the FMF block and the TRL losses, $\mathcal{L}_{TR}$ and $\mathcal{L}_{Reg}$. Next, we add the intermediate dynamic frames to obtain the mix configuration, without FMF, and lastly, we add our FMF block. We clearly observe the impact of the FMF block, especially on the OOD benchmark, where the AUC increases from 41.8\% to 43.7\%.
Next, \Cref{fig:all_ablation}(middle) shows the ablation on the TRL losses, $\mathcal{L}_{TR}$ and $\mathcal{L}_{Reg}$. The addition of either of them improves performance; however, the improvement is more significant when used together.
Further, in \Cref{tab:att_comp} we test the impact of FMF on accuracy compared to PSA\cite{liu2021polarized}, observing no noticeable fluctuations in performance.
We further evaluate the squeeze rate of our FMF block in \Cref{fig:all_ablation}(bottom). There we notice performance degradation on the OOD benchmark when we do not squeeze the FMF module, but not as much in the ID benchmark, suggesting that squeeze is more important for accurate filtration of OOD sequences.
In \Cref{tab:component_ablation_updates}, we show a case-study on parameter-free dynamic update strategies, where ours consistently improves over the others.

\section{Conclusions} \label{sec:conclusion}

We introduce SiamABC, a new Siamese visual tracker that improves the trade-off between the computational requirements and the OOD generalization ability, thus expanding the horizon of applicability of visual trackers in-the-wild under resource constraints. We have shown that it can be as fast as FEAR-XS, while being significantly more accurate with an evaluation over 11 benchmarks. We have also shown the superior ability of SiamABC in OOD generalization by reaching near-SOTA accuracies on the challenging OOD benchmark AVisT, with a significant improvement over the efficient Transformer-based SOTA methods. We credit this achievement to the four major technical contributions of the approach that include the use of a dual-search-region, the fast filtration layer FMT, the TRL loss, and the introduction, for the first time, of the dynamic TTA during tracking. Promising future extensions of this work may include further development of TTA for tracking, and the adoption of tracking inertia.


\section*{Acknowledgments}

Research reported in this publication was supported by the National
Institute Of Mental Health of the National Institutes of Health under
Award Number R44MH125238. The content is solely the responsibility of
the authors and does not necessarily represent the official views of
the NIH. This material is also based upon work supported by the
National Science Foundation under Grants No. 1920920, 2223793.

{\small
\bibliographystyle{ieee_fullname}
\bibliography{main}
}
\clearpage
\appendix

\section{Further Details on Methods}
\subsection{Fast Mixed Filtration}

We provide more details regarding the operations performed by the FMF block. Please refer to \Cref{fig:attention2} for a schematic representation of FMT. When fusing the pair of feature maps coming either from the dual-template or the dual-search-region, we first concatenate the two feature maps resulting in a representation with $C+C=2C$ channels dimension. Here, we aim to enhance the most relevant features through this filtration process. We further want to compress the feature space from $2C \times H \times W \rightarrow C \times H \times W$; therefore, we choose a squeeze rate of $S = 2$. Here, $W^V$ and $W^Q_{sp}$ perform the squeeze for $V$ and $Q_{sp}$, respectively, for the channel and spatial filters, where an average pooling operation is applied to $W^Q_{sp}$ to decrease the spatial dimensions of $W^Q_{sp}$ from $C \times H \times W \rightarrow C \times 1 \times 1$.  Moreover, $W^Q_{ch}$ decreases the channel dimensions of $Q_{ch}$ from $2C \times H \times W \rightarrow 1 \times H \times W$. A softmax function, $\phi$, is applied to $W^Q_{ch}$ and $W^Q_{sp}$ to produce the channel and spatial filter masks, respectively. Also, broadcast element-wise multiplications, $\star$, and element-wise summation, $\sum$, are then performed between $W^V$ and $\phi(W^Q_{ch})$ and between $\phi (W^Q_{sp})$ and $W^V$ to produce channel and spatial filters of size $C \times 1 \times 1$ and $1 \times H \times W$, respectively. Since there was a squeeze operation on the channel filter, $W_{ch}$ is applied to the channel filter to unsqueeze the channel dimensions from $C \times 1 \times 1 \rightarrow 2C \times 1 \times 1$, followed by a Layer Normalization layer to normalize the feature distributions. Furthermore, a sigmoid operation, $\sigma$, is performed on these filters, and then a broadcast element-wise summation, $\oplus$, is performed to form a total filter map of size $2C \times H \times W$. An element-wise multiplication operation is performed between the full filter map and $x$ to enhance the most important features from $x$ to form the final output $\bold{x}$. After this mixed-filtration of $x$, we further readjust the number of channels from  $2C \times H \times W \rightarrow C \times H \times W$, giving us $\bold{\check{x} }$, to perform future correlations.

Finally, like in~\cite{Huang2021}, the classification and regression bounding box heads have been implemented with separable convolutions.

\begin{algorithm}[t!]
  \centering
  \caption{Sampling Strategy}\label{alg:sampling}
  \begin{algorithmic}
  \State $X \gets rand(X^n) | X \in X^n$ \Comment{sequence}
  \State $I_T \gets x_i |  x_i \in X$ \Comment{static template}
  \State $I_t \gets x_j | x_j \in X[i, i+\Delta]$ \Comment{search-region} 
  \State $I_S \gets x_k, x_k \in X[i, j]$ \Comment{dynamic-search-region}
  \State $I_D \gets crop(I_S)$ \Comment{dynamic template}
  \end{algorithmic}
\end{algorithm}

\subsection{Training Sampling Strategy} \label{sec:methods_sampling}

We sample a random sequence from a database as shown in \Cref{alg:sampling}. We choose a random frame for the template, $I_T$. Then, we choose another frame randomly for the search-region, $I_t$, from an interval of size $\Delta=150$ right after the frame that contains the template. The dynamic-search-region $I_S$, and $I_D$, are randomly sampled from the interval with the lower bound of the template frame and the upper bound of the search-region frame. Here, we should note that $I_S$ contains $I_D$. We randomly choose 300,000 samples from GOT-10k \cite{Huang2021}, 100,000 from LaSOT \cite{fan2021lasot}, 200,000 from COCO2017 \cite{lin2014microsoft}, and 400,000 from TrackingNet \cite{muller2018trackingnet} datasets to collect a total of $10^6$ samples every epoch.

\subsection{Dynamic Sample Update Strategy}
\Cref{alg:dynamic_update} describes the details of our dynamic sample update strategy with $N=60$ and ${\lambda}_{D} = 0.25$. This strategy is parameter-less and lightweight.

\begin{algorithm}[th]
  \centering
  \caption{Dynamic Sample Update}\label{alg:dynamic_update}
  \begin{algorithmic}
    \State $t \gets 0$
    \State $x_t \gets $ sequence frame at time $t$
    \State $I_T \gets x_t$, $I_S \gets x_t$, $I_D \gets x_t$
    \State $init\_tracker(I_T)$ \Comment{initialize the tracker}
    \State $set\_dyn(I_S, I_D)$ \Comment{set the dynmaic samples}
    \State $C \gets 0$ \Comment{counter}
    \State $\overline{\rho}_t \gets 1$ \Comment{initialize average classification score}
  
    \While{$x_{t+1}$ is available}
      \State $C \gets C+1$ \Comment{increase the counter}
      \State $I_t \gets x_{t+1}$ \Comment{search-region} 
      \State ${\beta}_t, {\rho}_t \gets tracker(I_t)$ \Comment{ $\beta \gets$ bounding box}
      \State \Comment{$\rho \gets$ classification score}
      \If{$C \geq N$ \& ${\rho}_t > \overline{\rho }_{t}$}  \\ \Comment{$N \gets$ hyperparameter}
            \State $I_S, I_D \gets update(x_{t+1}, {\beta}_t)$ %
            \State $reset\_dyn(I_S, I_D)$ \\ \Comment{reset the dynmaic samples}
            \State $C \gets 0$
      \EndIf 
      \State $\overline{\rho }_t = (1-{\lambda}_{D}) \overline{\rho}_{t} +  {\lambda}_{D} {\rho}_t$ \\ \Comment{${\lambda}_{D} \gets$ hyperparameter}
      \EndWhile
  \end{algorithmic}
\end{algorithm}

\section{Additional Experiments}

\subsection{Evaluating Benchmarks}
We explain the strategies to evaluate our approach against other trackers. The FPS numbers were reproduced using the codebase provided by LightTrack \cite{yan2021lighttrack}. We follow GOT-10k\cite{Huang2021} guidelines and use their toolkit \cite{githubGitHubGot10ktoolkit} to evaluate the generic benchmarks. For VOT2020 \cite{kristan2020eighth}, we use their provided toolkit for the challenge. For GOT-10k\cite{Huang2021} and TrackingNet \cite{muller2018trackingnet}, we send the raw results to their servers for a fair evaluation. 

\subsection{Test-Time Adaptation Baselines}
There are no specific Test-Time Adaptation (TTA) approaches available for single-object tracking. Therefore, to compare our Dynamic TTA strategy mentioned in \Cref{sec:methods_tta}, we implement relevant backward-free and backward-based baselines according to the codebase provided by \cite{alfarra2023revisiting}. We carefully follow the instructions provided by the respective approaches and use them under our online setting. Here, our approach continually adapts to the new domains without forgetting as the BN-statistics stay anchored to the source statistics and model parameters remain frozen. Additionally, our approach continually adapts to each new instance with negligible difference in latency.

\subsection{Qualitative Assessment}
\begin{figure*}[t!]
  \hspace*{-5pt}
    \centering
     \includegraphics[width=18cm]{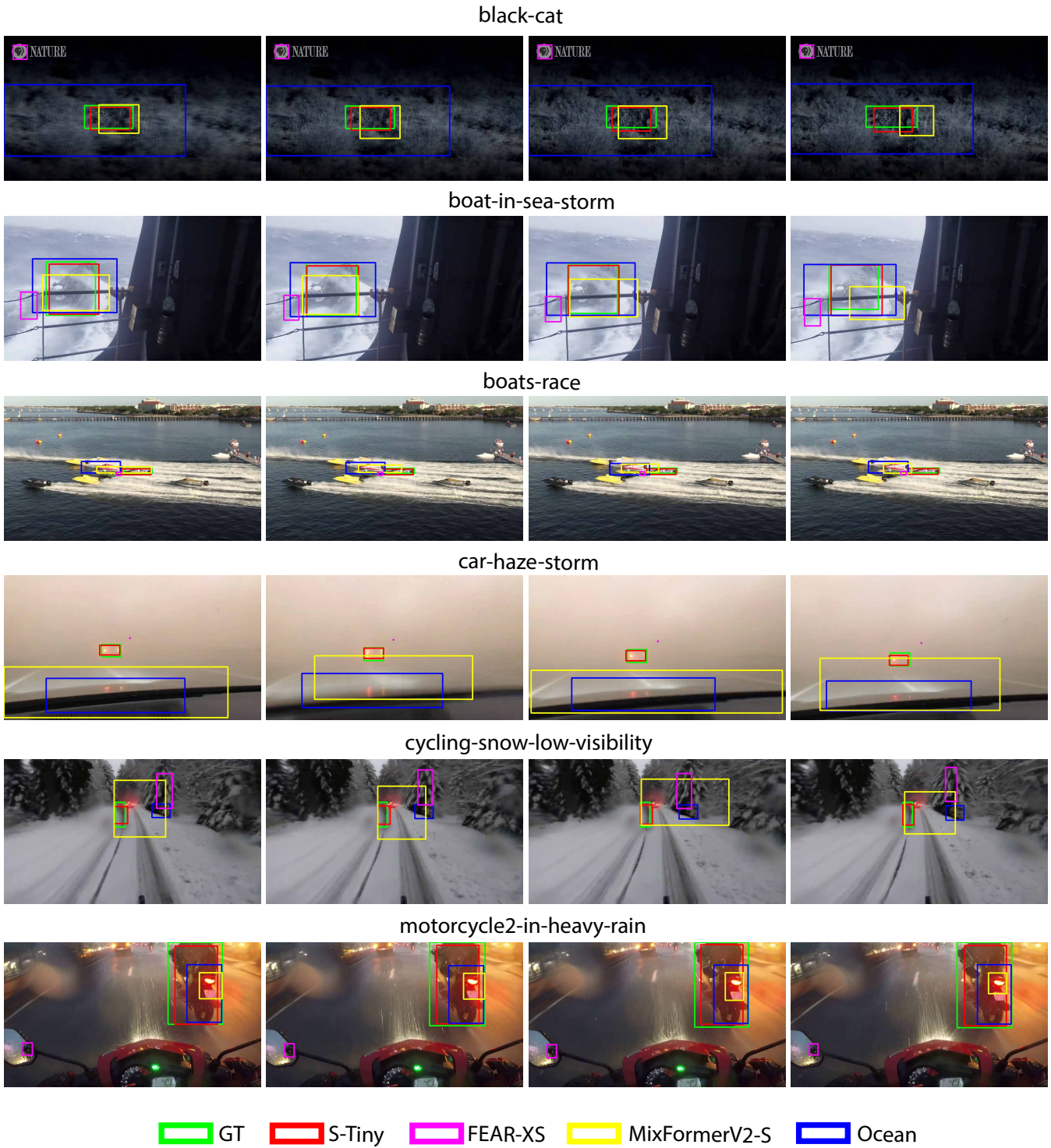}
     \caption{Qualitative comparison on the AVisT \cite{noman2022avist} dataset with other efficient trackers, and with the further inclusion of Ocean. Under adverse visibility conditions, our tracker, S-Tiny, is relatively stable compared to the others while running at 100 FPS on a CPU.}
     \label{fig:qual_compare_bbox}
\end{figure*}

We show qualitative results on the AviST~\cite{noman2022avist} benchmark, comparing S-Tiny with other trackers in \Cref{fig:qual_compare_bbox}. S-Tiny shows remarkable resilience against adverse visibility conditions. We also show qualitative results regarding the ablation assessment of the FMF layer compared to the baseline and the naive concatenation of features in \Cref{fig:qual_compare_bbox_abl}, visually showing the significance of our FMF layer.

\section{Potential Negative Societal Impact}
Our goal is mainly to advance the tracking performance of efficient visual tracking in-the-wild. On the other hand, if misused or used with ill-intent, this technology could potentially raise privacy concerns, especially if used for surveillance purposes, or other illegal activities like for stalking, bullying, etc. However, these are general issues that may arise even with other trackers that have been developed in the past. Our hope is to advance technology, and with regulations on its applications, we believe it should be possible to prevent its misuse.

\begin{figure*}[t!]
  \hspace*{-5pt}
    \centering
     \includegraphics[width=18cm]{figures/qaual_compare_bbox_abl.pdf}
     \caption{ Qualitative results on the ablation study of the FMF layer on the AVisT \cite{noman2022avist} dataset. We notice the higher stability of our tracker when using FMF in terms of overalps with the ground truth (green), in comparison to the baseline and the naive concatenation.}
     \label{fig:qual_compare_bbox_abl}
\end{figure*}


\end{document}